\documentclass[10pt,twocolumn,letterpaper]{article}

\usepackage{wacv}
\usepackage{times}
\usepackage{epsfig}
\usepackage{graphicx}
\usepackage{amsmath}
\usepackage{amssymb}
\usepackage{booktabs}
\usepackage{float}
\usepackage[dvipsnames]{xcolor}

\def\wacvPaperID{403} 

\wacvalgorithmstrack   

\wacvfinalcopy 

\ifwacvfinal
\usepackage[breaklinks=true,bookmarks=false]{hyperref}
\else
\usepackage[pagebackref=true,breaklinks=true,colorlinks,bookmarks=false]{hyperref}
\fi

\begin{document}

\title{BoxMask: Revisiting Bounding Box Supervision for Video Object Detection}

\author{
Khurram Azeem Hashmi
\and
Alain Pagani
\and
Didier Stricker
\and
Muhammamd Zeshan Afzal
\\
DFKI - German Research Center for Artificial Intelligence \\
{\tt\small firstname[0]\_firstname[1].lastname@dfki.de}
}


\maketitle
\thispagestyle{empty}

\begin{abstract}
    We present a new, simple yet effective approach to uplift video object detection. We observe that prior works operate on instance-level feature aggregation that imminently neglects the refined pixel-level representation, resulting in confusion among objects sharing similar appearance or motion characteristics. To address this limitation, we propose BoxMask, which effectively learns discriminative representations by incorporating class-aware pixel-level information. We simply consider bounding box-level annotations as a coarse mask for each object to supervise our method. The proposed module can be effortlessly integrated into any region-based detector to boost detection. Extensive experiments on ImageNet VID and EPIC KITCHENS datasets demonstrate consistent and significant improvement when we plug our BoxMask module into numerous recent state-of-the-art methods. 
\end{abstract}

\section{Introduction}
\label{sec:intorduction}
With the recent advancements in deep convolutional neural networks~\cite{he2016deep, xie2017aggregated, wang2020deep}, object detection in still images has gained a remarkable progress~\cite{girshick2014rich,ren2015faster,redmon2016you,sun2021sparse, ge2021yolox}. The naive idea of applying image-based detectors on each frame to perform Video Object Detection (VOD) often underperforms, owing to the deteriorated object appearance due to motion blur, rare poses, and part occlusions in videos. Therefore, exploiting the encoded temporal information in videos~\cite{zhu2017flow, zhu2017deep, wu2019sequence, gong2021temporal} has become a de facto choice to tackle these challenges. 

Earlier video object detection techniques utilizing temporal information mainly operate under two paradigms. The first category of methods applies post-processing on temporal information to make still image object detection results~\cite{han2016seq, kang2016object, kang2017t, belhassen2019improving} more consistent and stable. Alternatively, the second group leverages the feature aggregation of temporal information~\cite{zhu2017flow, chen2018optimizing, wu2019sequence, yao2020video, chen2020memory, gong2021temporal}. Albeit these region-based state-of-the-art systems have greatly boosted the performance of VOD, they suffer from differentiating the confusing objects with similar appearances or uniform motion attributes.  

\begin{figure*}
\begin{center}
    \includegraphics[width=14 cm]{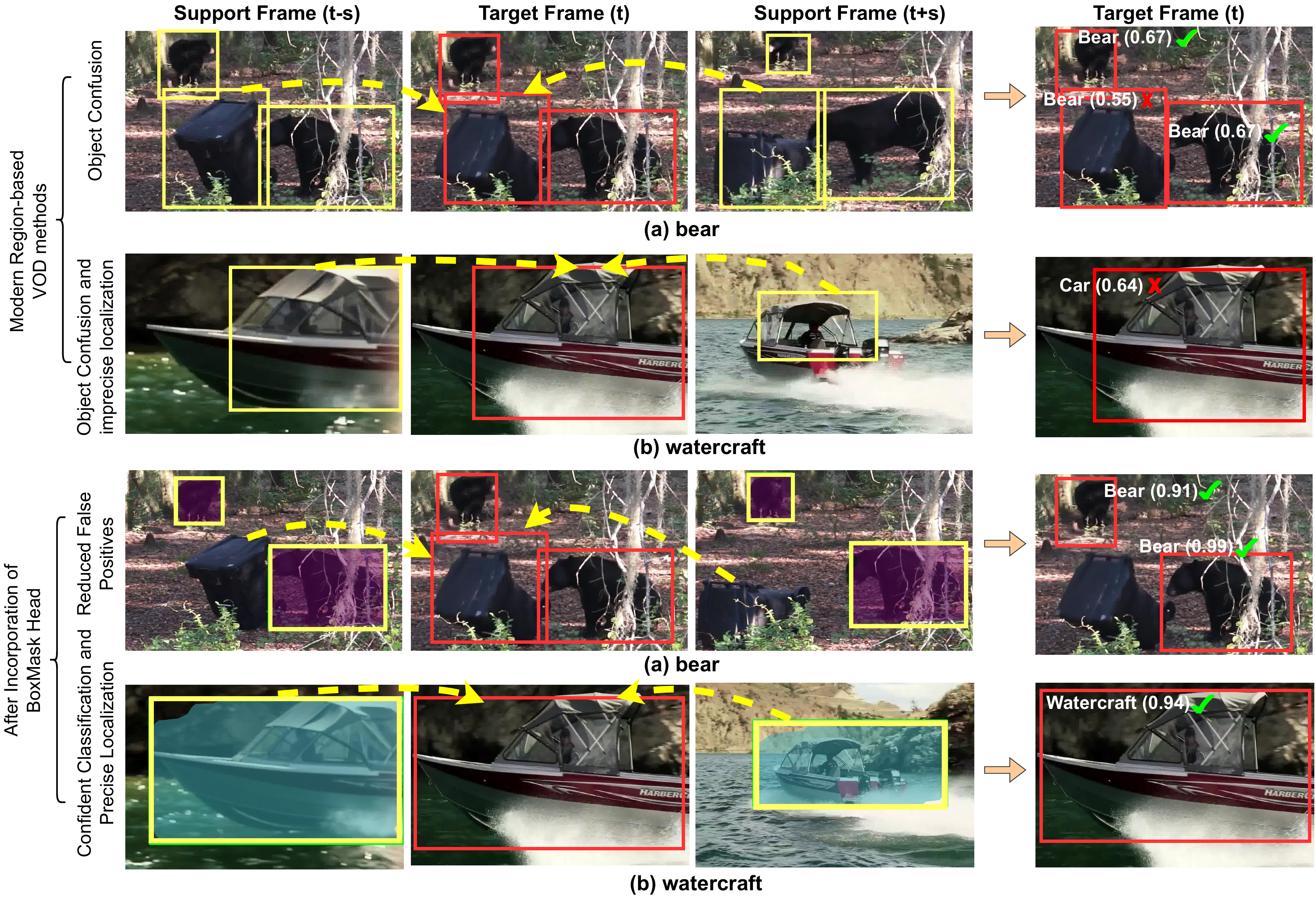}
    \caption{Motivation.~Despite leveraging spatio-temporal information from support frames $t-s$ and $t+s$, modern VOD methods often misclassify objects with similar appearance and uniform motion characteristics.~For instance, a moving object in the {background} is categorized as a \textit{bear} in (a), while \textit{Watercraft} is mistaken for a \textit{Car} in (b). To address this, we devise a simple BoxMask module that learns pixel-level features by introducing crucial discriminative cues to boost detection among confused object categories. Note that with fine-grained pixel-level learning, our BoxMask removes misclassification of {background} in (a) and correctly categorizes \textit{Watercraft} in (b). Best view it on the screen.}
\end{center}
\label{fig:prob_def}
\vspace{-20pt}
\end{figure*}

We observe that most of the previous approaches~\cite{zhu2017flow, wu2019sequence, gong2021temporal, chen2020memory} operate on instance-level feature aggregation that imminently neglects the refined pixel-level representation, resulting in acceptable localization but inferior classification. As illustrated in the first two rows of Figure~\ref{fig:prob_def}, although the object detector exploits spatio-temporal context from support frames (${t -s}$ and ${t + s}$) to refine proposal features, it produces false positives by classifying background as a \textit{Bear} and misclassifies \textit{Watercraft} with a \textit{Car} at the target frame \textit{t}. To overcome this hurdle, we design a novel module called BoxMask that exploits class-aware pixel-level temporal information to boost VOD. Inspired by~\cite{he2017mask} in still images, the BoxMask predicts a class-aware segmentation mask for each region of interest along with the conventional classification and localization. Since this paper deals with the problem of object detection in videos, we investigate bounding box-level annotation to generate coarse masks which supervise our BoxMask network. The advantages of adopting our BoxMask head are two folds. First, the class-aware pixel-level features reduce the hard false positives between objects with low spatial and temporal inter-class variance. Second, since the size of the predicted mask is identical to the target region, fine-grained pixel-level learning assists the detector in precise localization.~We summarize the main contribution of this paper as follows:
\vspace{-5pt}
\begin{itemize}
    \item We observe that object misclassification 
    is the crucial obstacle that limits the upper bound of existing video object detection methods.~We further revisit the idea of leveraging bounding box annotations to supervise both regression and mask prediction (see Figure~\ref{fig:prob_def}).
    \vspace{-5pt}
    \item We propose BoxMask, an extremely simple yet effective module that learns additional discriminative representations by incorporating class-aware pixel-level information to boost VOD.
    \vspace{-5pt}
    \item Our BoxMask is a plug-and-play module and can be integrated into any region-based detection method. With our novel class-aware pixel-level learning introduced in recent state-of-the-art methods, we achieve an absolute gain of 1.8\% in mAP and 2.1\% in mAP on ImageNet VID and EPIC KITCHENS benchmarks, respectively.

\end{itemize}

\section{Related Work}
\label{sec:related_work}

\noindent \textbf{Object Detection in Images.}
\label{sec:rel_object_detection}
The existing methods in image-based object detection can be mainly divided into single-stage detectors~\cite{liu2016ssd,redmon2016you,redmon2017yolo9000, redmon2018yolov3, chen2021you, ge2021yolox} and multi-stage or region-based detectors~\cite{ren2015faster, cai2018cascade, chen2019hybrid, guo2020augfpn, jiang2020sp}. Mask R-CNN~\cite{he2017mask} replaces RoI Pooling with RoIAlign and introduces an extra instance segmentation head that not only improves instance segmentation but advances object detection. Cheng et al.~\cite{cheng2018revisiting} blame the weak classification head for inferior detections and propose to ensemble the classification scores of Faster R-CNN~\cite{ren2015faster} and R-CNN~\cite{girshick2014rich} as a remedy. IoU-Net~\cite{jiang2018acquisition} proposes a separate confidence mechanism for localization. Double-Head R-CNN~\cite{wu2020rethinking} disentangles the detection head by treating classification with the fully connected head and regression with a convolution head. Along with this direction, seminal work~\cite{song2020revisiting} incorporates TSD in a region-based detector~\cite{ren2015faster} that learns different features for classification and regression. Later, separate losses are added to the whole loss function to optimize detection. Similar to these works in still images~\cite{he2017mask, wu2020rethinking, song2020revisiting, jiang2018acquisition}, we observe that a naive sibling head in the region-based detector~\cite{ren2015faster} confuses objects with similar motion characteristics and leads to sub-optimal video object detection.

\noindent \textbf{Box-supervised Semantic and Instance Segmentation in Images.}
\label{sec:rel_bounding_box}
There has been an increasing trend in exploiting bounding box annotations to enhance weakly supervised instance and semantic segmentation approaches in still images~\cite{dai2015boxsup, bbox_2seg, bbox_SDI, bounding_box, budget_bbox}. The main reason is that bounding boxes contain knowledge about the precise location of each object, and they are approximately 35 times faster to annotate than per-pixel labeling~\cite{everingham2010pascal, bearman2016s}. Along with a similar direction, our work exploits box-level annotations to generate coarse masks, eventually boosting video object detection.

\begin{figure*}
\begin{center}
\includegraphics[width=10cm]{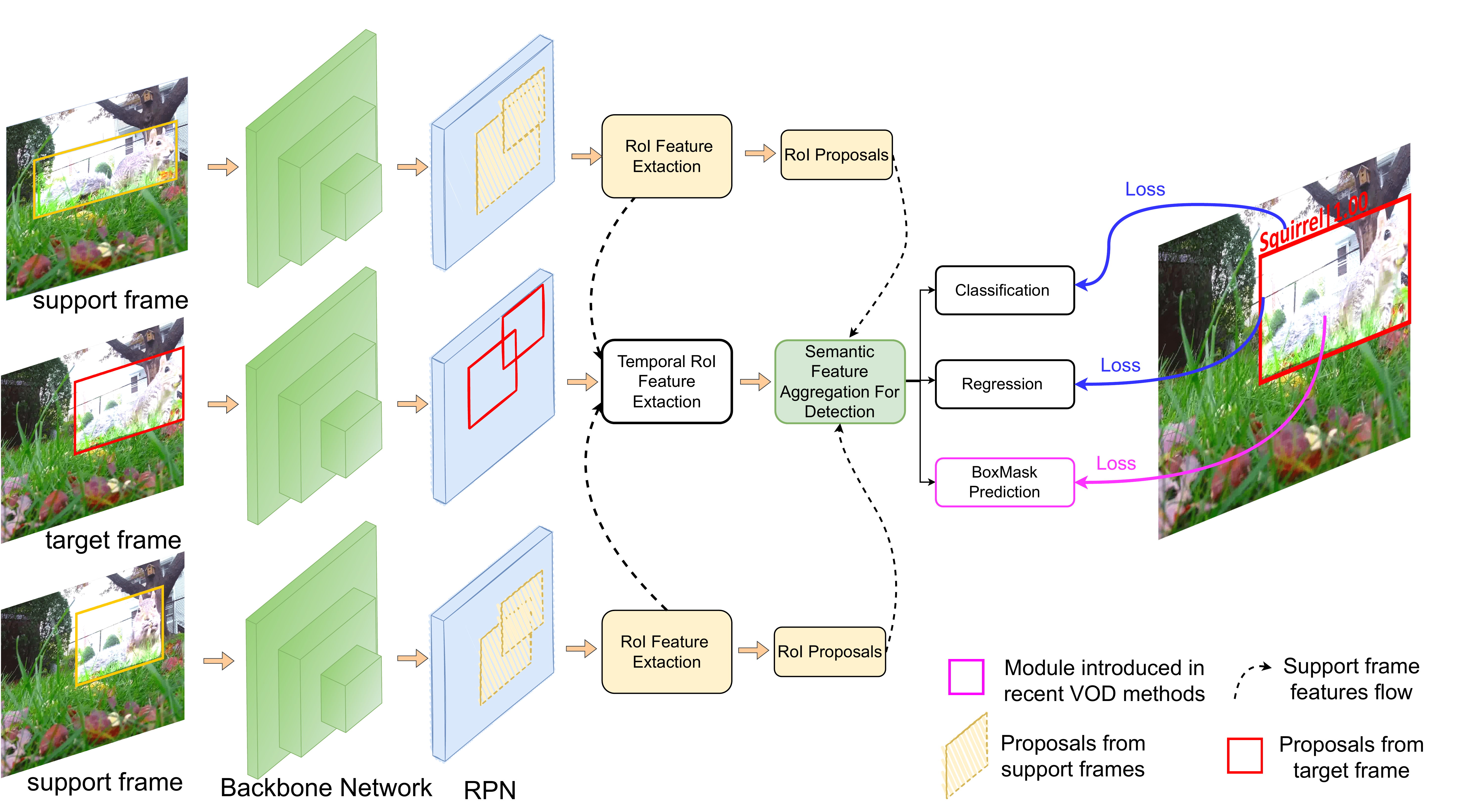}
\caption{Architectural overview of modern region-based VOD methods and our proposed modules highlighted in magenta. Alongside spatio-temporal features, our method introduces important class-aware pixel-level features, which effectively tackles object confusion to boost performance in modern region-based video object detection methods.}
\label{fig:pipeline}
\end{center}
\vspace{-20pt}
\end{figure*}

\vspace{3pt}
\noindent \textbf{Object Detection in Videos.}
\label{sec:rel_vod}
Prior methods for video object detection have two directions. One direction exploits the redundancy in video frames by incorporating optical flow~\cite{zhu2017deep, zhu2018towards}, scale-time lattice~\cite{chen2018optimizing}, reinforcement learning capabilities~\cite{yao2020video}, and heatmaps~\cite{xu2020centernet} to reduce the cost of the feature extraction process by propagating keyframe features to other frames in videos. Another line of work leverages temporal information encoded in videos to boost VOD, and our work operates on this trend. Existing techniques exploit temporal information in two ways. The first way is to refine the detection results with post-processing methods~\cite{han2016seq, kang2016object, kang2017t}. Although these approaches improve the performance of VOD, they heavily rely on the image-based detector trained with no knowledge of temporal information.
On the contrary, The second direction is to capitalize temporal information during the training stage~\cite{zhu2017flow, wang2018fully, feichtenhofer2017detect, bertasius2018object, guo2019progressive, shvets2019leveraging, wu2019sequence, deng2019relation, zhu2017deep, zhu2018towards, chen2018optimizing, xiao2018video, deng2019object, tang2019object,chen2020joint,chen2020memory, gong2021temporal, zhou2022transvod}. Some of these methods utilize optical flow~\cite{dosovitskiy2015flownet} to warp and aggregate features across frames~\cite{zhu2017flow, wang2018fully,kang2017t}. Despite the improvement, the optical flow based-methods fail in the case of occlusions. Most existing region-based VOD methods~\cite{zhu2017flow, zhu2017flow, wu2019sequence, gong2021temporal} tackle the inherent challenges by aggregating temporal features. However, they mainly rely on instance-level feature aggregation, which pays less attention to the content of object proposals, resulting in confusion between objects with similar appearance and motion characteristics.~Very recently, TransVOD~\cite{zhou2022transvod} introduces the transformer-based VOD method by extending the Deformable DETR~\cite{zhu2020deformable} with a temporal transformer to aggregate object queries from different video frames.

\noindent \textbf{Tackling Object Confusion in Videos.}
Han et al.~\cite{han2020mining} are the first to highlight object confusion as to the main problem in VOD. They propose exploiting inter-video and intra-video proposal relations to tackle object confusion. Another seminal works~\cite{han2020exploiting, han2021class} attempts to solve this problem by devising better feature aggregation schemes that enhance target frame feature representation. Despite the gratifying improvement in detection, these approaches rely on a region-based detector that focuses more on discriminating between background and foreground regions than differentiating between various foreground regions~\cite{cheng2018revisiting}. Moreover, these methods operate on complex pipelines to produce impressive results. Alternatively, we design a simple but effective BoxMask module that achieves similar performance upon integrating into recent region-based VOD methods.

\section{Method}
This section first describes an overview of the modern region-based detectors in VOD by diving into the inherent misclassification problem in Section~\ref{sec:region_based_vod}. Later, we explain the proposed BoxMask module and its learning mechanism in Sections~\ref{sec:boxMask} and~\ref{sec:boxMask_learning}, respectively.

\subsection{Revisiting Region-based Detectors in VOD}
\label{sec:region_based_vod}
Figure~\ref{fig:pipeline} depicts an overview of region-based detectors in VOD. First, a backbone network extracts spatial features from the target frame (the actual frame on which detection needs to be executed) and support frames (other video frames that assist the detection on a target frame). Subsequently, a Region Proposal Network (RPN)~\cite{ren2015faster} predicts object proposals for each frame and aims to minimize the regression loss $L_{reg}$ and classification loss $L_{cls}$ defined as:
\begin{equation}
\vspace{-2pt}
\label{eq:rpn_loss}
     L_{rpn} = L_{cls}(p,p^{*}) + p^*.L_{reg}(t, t^{*}) 
\end{equation}
where $p$ is the estimated probability of a proposal being an object and $p^*$ represents 1 or 0 depending upon the label of the anchor box. The term $t$ denotes the coordinates of the predicted object proposal, and $t^*$ is the ground truth. Here, note that the classification loss $L_{cls}$ in Equation~\ref{eq:rpn_loss} only focuses on improving the objectness of proposals instead of object classification.

\begin{figure}
\centering
\includegraphics[width=8.4cm]{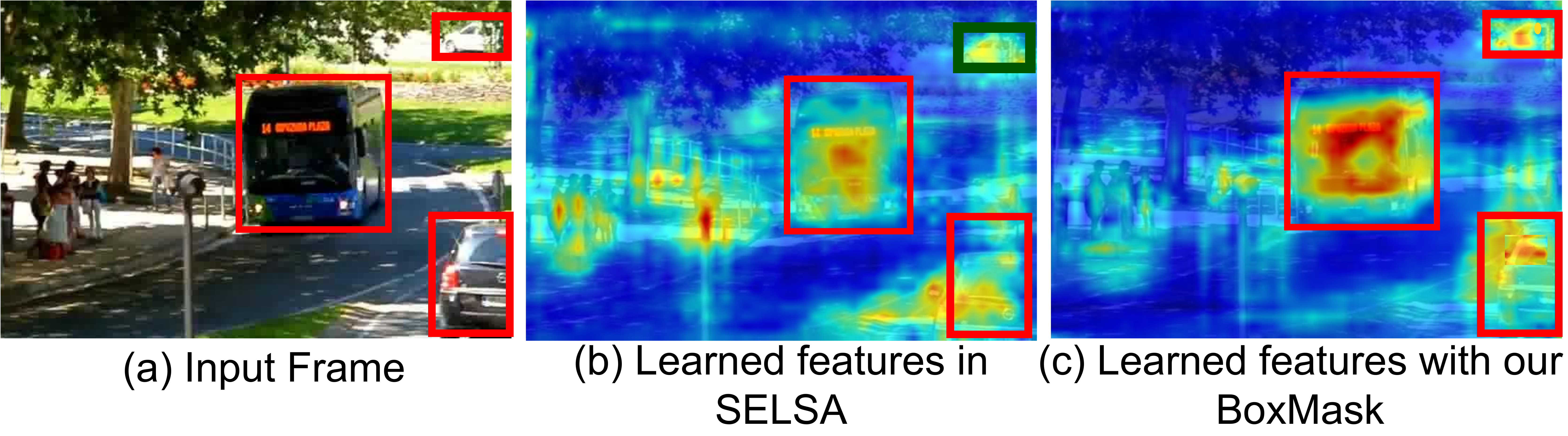}
\caption{Visualization of learned class activation maps of SELSA and SELSA+BoxMask. (a) shows a sample input frame with target bounding boxes in \color{red}{red}\color{black}. (b) highlights that existing instance-level aggregation methods like SELSA~\cite{wu2019sequence} imminently pay less attention to blurred and partly occluded objects, leading to missed detections (highlighted in \textcolor{ForestGreen}{green}\color{black}). (c) depicts that our fine-grained pixel-level learning brings additional discriminative cues that enrich target object features and suppress needless features.}
\label{fig:feature_vis}
\vspace{-10pt}
\end{figure}

In the second stage, feature aggregation is performed between object proposal features of the target frame and support frames in a video. These aggregated features are pooled by an RoI Align pooling operator and propagated to the detection head designed to optimize multi-class classification and regression. For training, the detection loss is given by:
\begin{equation}
\label{eq:rcnn_loss}
    L_{det} = L_{cls}(p_{c},y) + L_{reg}(t, t^{*}) 
\end{equation}
where $p_{c}$ represents the predicted class distribution and $y$ is the class label of an object in a target frame.~For comprehensive details about the parameterization of RPN and detection head, we refer readers to~\cite{ren2015faster}. Since the optimization of these region-based detectors relies on the cumulative sum ($L_{cls} + L_{reg}$), it converges to a compromising suboptimal of two tasks~\cite{cheng2018revisiting}.~Consequently, despite aggregating object proposal features from several support frames, the performance of current state-of-the-art VOD methods degrades due to the underlying object confusion caused by similar appearance and uniform motion characteristics. Furthermore, most existing methods operate on instance-level feature aggregation, which ignores blurred and partly occluded instances, leading to missed detection, as illustrated in Figure~\ref{fig:feature_vis}\color{red}{(b)}\color{black}. In this paper, we aim to alleviate these limitations by incorporating class-aware pixel-level information in the detection head that brings additional discriminative features. To this end, we propose BoxMask, which assists the optimization of the detection head by enhancing target object features and discouraging irrelevant features, as visualized in Figure~\ref{fig:feature_vis}\color{red}{(c)}\color{black}.

\subsection{BoxMask}
\label{sec:boxMask}
\begin{figure}
\centering
\includegraphics[width=8.5cm]{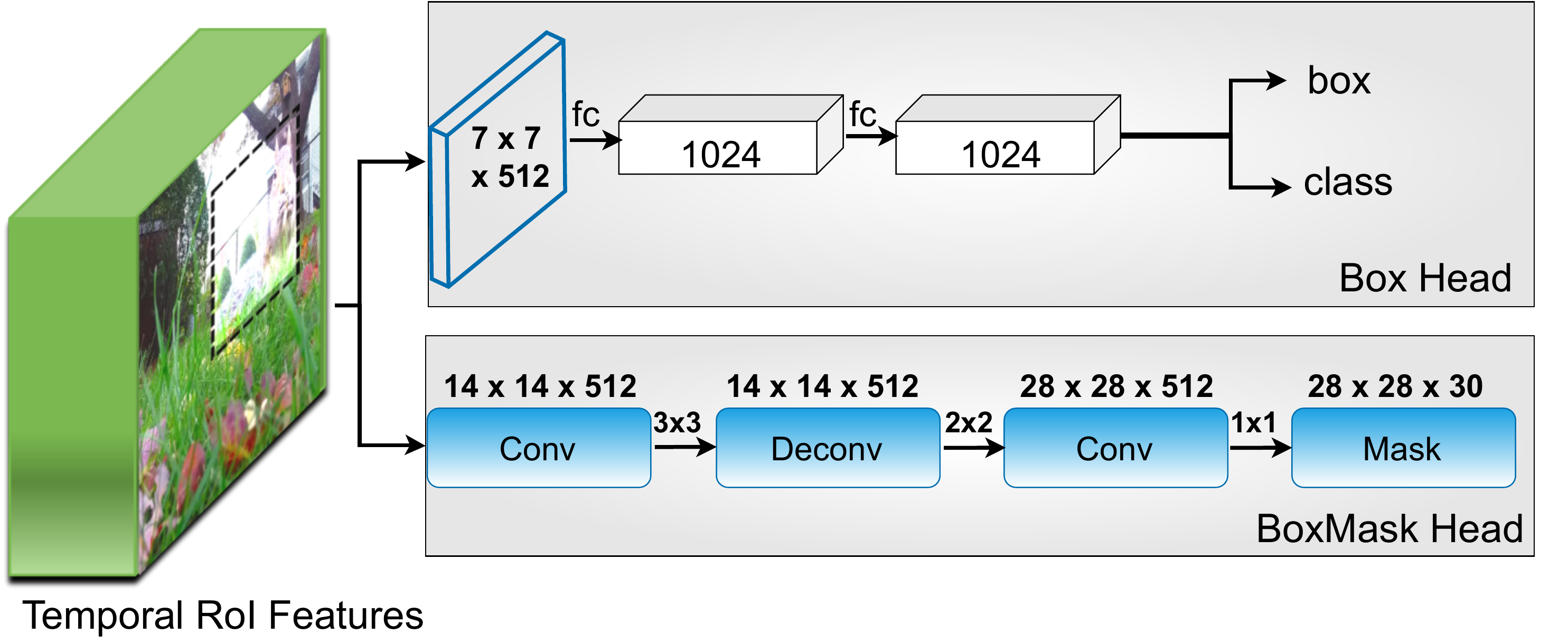}
\caption{This figure presents the overall architecture of the detection phase equipped with a BoxMask head at the bottom. Numbers on blocks represent spatial resolution and channels, whereas numbers on arrows are the size of the kernel. In the BoxMask head, Conv and Deconv denote convolutions and deconvolutions, respectively. Thanks to its simplistic design, the proposed BoxMask head can be integrated into any region-based VOD method.}
\label{fig:boxmask_detection}
\vspace{-10pt}
\end{figure}

After extracting object proposals by RPN, in the second stage, we also have a set of aggregated proposal features $O = \{o_k\}_{k=1}^{K}$ from target and support frames, where K is the number of proposals. The BoxMask head predicts a binary mask for each RoI for classification and regression. Note that contrary to predicting instance mask in Mask R-CNN~\cite{he2017mask}, our method predicts the mask of a complete bounding box along with classification and localization to simplify the overall multi-stage pipeline. Figure~\ref{fig:boxmask_detection} visualizes the integration of the BoxMask module in region-based VOD methods.

\vspace{3pt}
\noindent \textbf{Temporal RoI Feature Extraction for BoxMask Head.}
\label{sec:roi_feature}
RoIAlign~\cite{he2017mask} pooling has outsmarted the RoIPool~\cite{girshick2015fast} operation to extract feature maps for each RoI, and it has been widely used in recent state-of-the-art VOD methods~\cite{wu2019sequence, han2020mining}. Instead of the conventional RoIAlign operation, our method follows the spirit of~\cite{gong2021temporal} and exploits temporal information to extract RoI features.~Given a group of video frames $\{V_{t+s}\}_{s=-N/2}^{N/2}$ and corresponding feature maps$\{F_{t+s}\}_{s=-N/2}^{N/2}$ generated from the backbone network, where $F_{t}$ represents feature maps of the target frame, and $F_{t+s}$ ($s\neq0$) denotes feature maps of support frames.~First, we extract RoI features of target frame $R_{t}$ by applying conventional RoIAlign on the target frame proposals, and target frame feature maps $F_{t}$. Then, in order to extract the most similar support frame RoI features~$R_{t+s}$ for target frame RoI features $R_{t}$, we compute the cosine similarity $C_{t+s}\in \mathbb{R}^{H \times W}$ between support frame feature maps $F_{t+s}$ and a target frame RoI features $R_{t}$ as follows:
\begin{equation}
    \label{eq:cos_sim}
    C_{t+s} = R_{t} \otimes \{F_{t+s}\}^{T}
\end{equation}
where $\otimes$ represents matrix multiplication, and $\{.\}^{T}$ highlights the matrix transposition.~Later, analogous to~\cite{gong2021temporal}, we employ multi-head self-attention~\cite{vaswani2017attention} to aggregate target RoI features~$R_{t}$ and support frame RoI features~$R_{t+s}$ to form temporal RoI features for the target frame $\overline{R_{t}}$: 
\begin{equation}
    \overline{R_{t}} = \textit{MSA}(R_{t}, R_{t+s}) 
\end{equation}
where \textit{MSA} is a multi-head self-attention operation~\cite{vaswani2017attention}. An overview of temporal RoI feature extraction is depicted in Figure~\ref{fig:pipeline}. We refer readers to~\cite{gong2021temporal} for the detailed parameterization of temporal attentional feature aggregation of RoI features.

\noindent \textbf{Instance Feature Extraction and Prediction.}
The BoxMask head is a fully convolutional~\cite{long2015fully} instance segmentation head in which, first, the temporal RoIAlign operation extracts $14 \times 14$ RoI features that are propagated into a single $3 \times 3$ convolutional layer to learn instance features. Contrarily to the complex instance segmentation problem~\cite{he2017mask, chen2019hybrid}, we aim to predict the pixel mask of a rectangular bounding box. Therefore, we empirically establish that a single convolutional network is an optimal choice (see Section~\ref{sec:ablation_num_conv}). As depicted in Figure~\ref{fig:boxmask_detection}, our prediction head contains a $2 \times 2$ deconvolution with a stride of 2, followed by the $1 \times 1$ convolution that predicts an output mask of size $C.(m \times m)$ for each RoI, where C represents a total number of classes and $(m \times m)$ is the resolution.

\subsection{Learning and Optimization}
\label{sec:boxMask_learning}
To alleviate the problem of object confusion and imprecision localization in videos, we view detection as a pixel-level classification problem. Furthermore, since our method operates in an end-to-end manner, it is robust to various datasets and backbone networks.

\vspace{3pt}
\noindent \textbf{Generating Ground Truth. }
Considering our work deals with video object detection, an accurate object mask annotation is not available. Therefore, we revisit the exploitation of bounding boxes~\cite{dai2015boxsup, bounding_box, bbox_2seg} in VOD and generate a mask with the given bounding box annotations to supervise the BoxMask head. Given the ground truth of bounding boxes represented by $B_{box} \in \mathbb{R}^{K \times 5}$, where $K$ denotes the set of bounding boxes consisting of 4 coordinates along with a corresponding class label. We define the bounding box mask tensor as $M_{box} \in \mathbb{R}^{m \times (L+1)}$, where $m$ is the predicted spatial resolution and $(L+1)$ denotes L object classes and the background. We create a bounding box mask tensor $M_{box}$ by labeling all the pixels inside the box with a corresponding class label. Following~\cite{bbox_SDI, bbox_2seg}, if two boxes overlap, we consider that a smaller box is in the front and label the pixels with a class of the smaller bounding box. The remaining pixels, not packed in any bounding box, belong to the background class.

\vspace{3pt}
\noindent \textbf{BoxMask Loss and Multi-Task Learning.}
We employ bounding box mask tensor $M_{box}$ to optimize the mask prediction by minimizing the cross-entropy loss $L_{bm}$:
\begin{equation}
\label{eq:ce_loss}
    L_{bm} = -\frac{1}{m} \sum_{c=0}^{L}\sum_{i=1}^{m}M(i,c)\log (y(i,c))
\end{equation}
where $L_{bm}$ allows the network to predict the class for each pixel in each sampled RoI. This decouples the prediction of mask and class labels. Moreover, it assists the feature learning for localization because the predicted mask aims to be proportionally identical to the target bounding box. Upon integrating the BoxMask head in region-based video object detection methods, the detection loss explained in Equation~\ref{eq:rcnn_loss} becomes
\begin{equation}
\label{eq:loss_function}
    L_{det} = L_{cls} + L_{reg} + \lambda L_{bm} 
\end{equation}
where $\lambda$ is the hyperparameter to control the weight of the BoxMask loss. We empirically set $\lambda$=0.5 in all experiments unless stated otherwise (refer to Section 2 in supplementary materials). 

\section{Experiments and Results}

\subsection{Experimental Setup}

\vspace{3pt}
\noindent \textbf{Datasets and Evaluation Metrics.}
We perform extensive experiments on the ImageNet VID dataset~\cite{russakovsky2015imagenet}. The dataset comprises 3862 training videos and 555 validation videos with labeled bounding boxes of 30 classes. Following existing methods~\cite{wu2019sequence, gong2021temporal, zhu2017flow, zhu2017deep}, we train our models on the intersection of ImageNet DET and VID datasets\cite{russakovsky2015imagenet} by utilizing the split provided in~\cite{zhu2017flow}. For direct comparison with prior works, we validate our models on the ImageNet VID validation set by using mean average precision (mAP) as a metric. 

\vspace{3pt}
\noindent \textbf{Training and Inference Details.}
We employ ResNet-50~\cite{he2016deep} as the backbone network for ablation studies. In addition to ResNet-50, we utilize more powerful ResNet-101~\cite{he2016deep} and ResNeXt-101~\cite{xie2017aggregated} to compare performance with existing methods. The backbone networks are initialized with ImageNet~\cite{krizhevsky2012imagenet} pre-trained weights. We use SGD to train our models on 7 epochs with a total batch size of 8 on 8 GPUs. The training starts with an initial learning rate of 0.01, which is divided by 10 at the 4--th and 6--th epoch. For direct comparison, we sample one training frame (target frame) and two random frames (support frames) from the same video. During inference, we sample T frames (support frames) from the same video in addition to the target frame. Adopting~\cite{bertasius2018object, gong2021temporal}, we replicate the first/last frame of the video if support frames exceed the video start/end. Since our method detects objects in a target frame, the BoxMask module is switched off during the inference. Analogous to prior works~\cite{gong2021temporal,wu2019sequence, zhu2017flow},~Non-Maximum Suppression~(NMS)~with an IoU threshold of 0.5 is incorporated to reduce reduplicate detections. The frames are resized to a shorter side of 600 pixels during both training and inference. For a detailed summarization of network architecture, refer to supplementary material (Section 1).

\subsection{Effect of BoxMask on ImageNet VID Benchmarks}
We compare performance between state-of-the-art systems equipped with our BoxMask module and summarize the results in Table~\ref{table:sota_comparison}.~For a fair comparison, we reproduce the results of recent methods~\cite{chen2020memory, cui2021tf, gong2021temporal, wu2019sequence, zhu2017flow} by utilizing the original code from the authors.~Therefore, for TF-Blender~\cite{cui2021tf}, we include results with their module crafted in FGFA~\cite{zhu2017flow}.~Looking at the results in Table~\ref{table:sota_comparison}, our proposed BoxMask brings consistent and significant gains when incorporated into existing state-of-the-art methods with all three backbones.~When BoxMask is plugged into TROI~\cite{gong2021temporal}, we accomplish new state-of-the-art results with 80.7\% mAP on the ResNet-50 backbone. Furthermore, leveraging our BoxMask module, all methods~\cite{chen2020memory, cui2021tf, gong2021temporal, wu2019sequence, zhu2017flow} with similar backbones enjoy gains from 0.4\%~(ResNeXt-101) to 1.8\%~(ResNet-50) in mAP. We argue that prior feature aggregation methods heavily rely on the capabilities of backbone networks, which results in inferior performance on a relatively weaker backbone of ResNet-50.~Alternatively, our pixel-level feature information in the BoxMask complements existing temporal feature aggregation schemes in~\cite{wu2019sequence, gong2021temporal}, obtaining superior gains in performance.

\begin{table}
\begin{center}
    \small
    \begin{tabular}{llll}
    \toprule\noalign{\smallskip}
      \textbf{Methods} &  \textbf{mAP}($\%$) &  \textbf{mAP}($\%$) &  \textbf{mAP}($\%$)\\ &\textbf{R-50} &\textbf{R-101} &\textbf{RX-101} \\
    \noalign{\smallskip}
    \hline
    SFB\textsubscript{NIPS'15}\cite{ren2015faster}  & 70.1 & 74.1 & 76.4 \\
    FGFA\textsubscript{ICCV'17}\cite{zhu2017flow}  & 74.7& 77.8 & 79.6 \\
    SELSA\textsubscript{ICCV'19}\cite{wu2019sequence}  & 78.4 & 80.2 &  83.1\\
    MEGA\textsubscript{CVPR'20}\cite{chen2020memory}  & 77.3 & 81.6 & - \\
    TF-Blender\textsubscript{ICCV'21}\cite{cui2021tf}  & 75.4 & 79.3 &  80.1\\
    TROI\textsubscript{AAAI'21}\cite{gong2021temporal}  &78.9 & 82.0 & \textbf{\color{blue}84.3}\\
    \hline
    
    \textbf{SFB + BoxMask}& 71.2$_{\uparrow}$\textsubscript{1.1} & 75.0$_{\uparrow}$\textsubscript{0.9} & 77.2$_{\uparrow}$\textsubscript{0.8} \\
    \textbf{FGFA + BoxMask }& 75.6$_{\uparrow}$\textsubscript{0.9} & 78.7$_{\uparrow}$\textsubscript{0.7} & 80.0$_{\uparrow}$\textsubscript{0.4}\\
    \textbf{SELSA + BoxMask }& \textbf{\color{blue}79.5$_{\uparrow}$\textsubscript{1.1}} & 81.1$_{\uparrow}$\textsubscript{0.9} & 83.5$_{\uparrow}$\textsubscript{0.4}\\
    \textbf{MEGA + BoxMask }& 78.2$_{\uparrow}$\textsubscript{0.9}& \textbf{\color{blue}82.3$_{\uparrow}$\textsubscript{0.7}} & - \\
    \textbf{TF-Blender + BoxMask }& 76.3$_{\uparrow}$\textsubscript{0.9} & 79.9$_{\uparrow}$\textsubscript{0.6} & 80.4$_{\uparrow}$\textsubscript{0.3}\\
    \textbf{TROI + BoxMask }& \textbf{\color{red}80.7$_{\uparrow}$\textsubscript{1.8}}& \textbf{\color{red}83.2$_{\uparrow}$\textsubscript{1.2}} & \textbf{\color{red}84.8$_{\uparrow}$\textsubscript{0.5}}\\
    
    \bottomrule
    \end{tabular}
    \caption{\color{black} Comparison with existing state-of-the-art methods on the ImageNet VID dataset. The SFB represents Single-Frame Baseline, Faster R-CNN, utilized as a base detector in all experiments.~R and RX denote ResNet and ResNeXt backbone networks. The two best results are highlighted in \color{red}{red} \color{black} and \color{blue}{blue}\color{black}.}
    \label{table:sota_comparison}
\end{center}
\vspace{-20pt}
\end{table}

\subsection{Qualitative Analysis}

\begin{figure}
\vspace{-10 pt}
\begin{center}
\includegraphics[width=8.5cm]{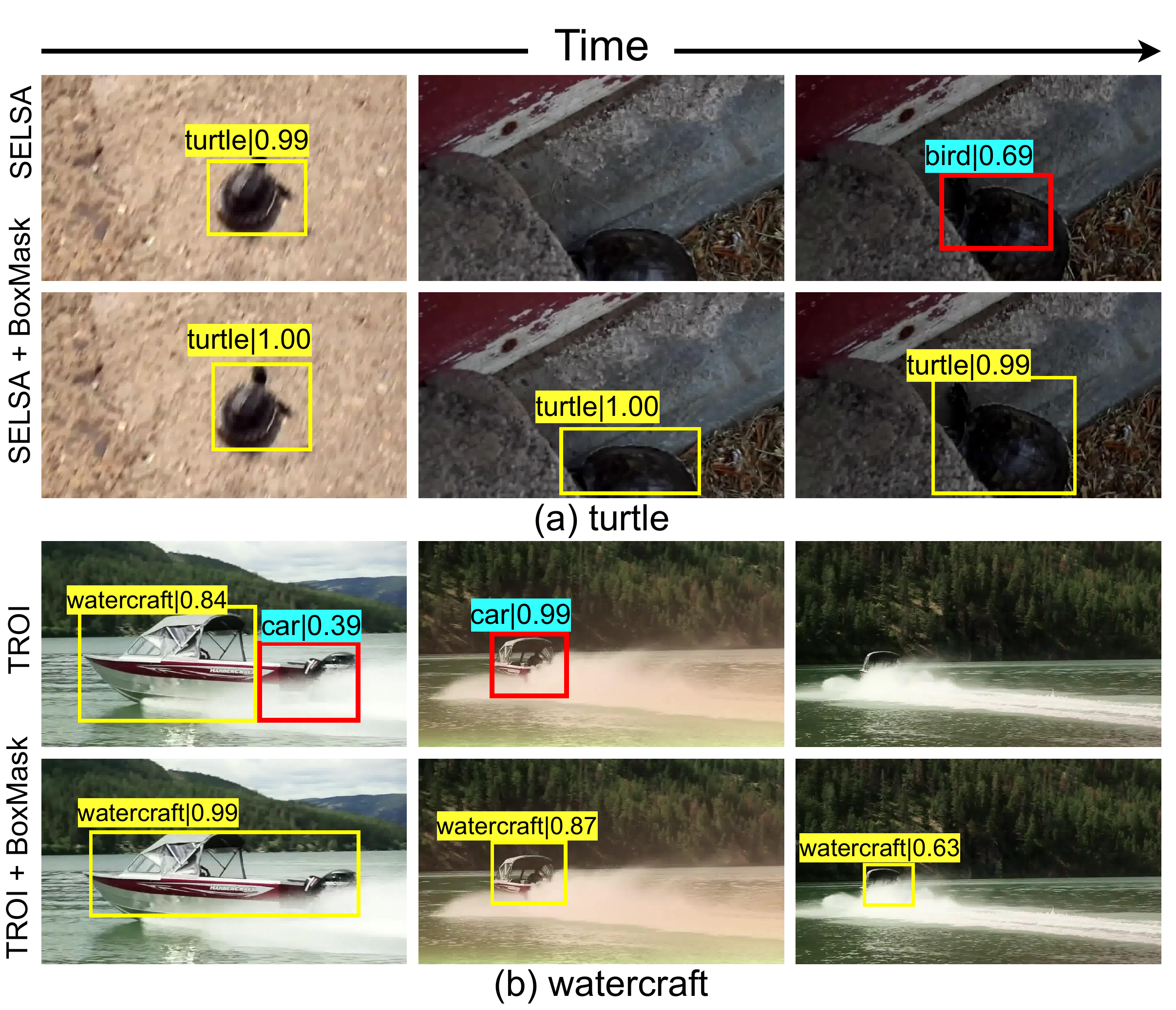}
\caption{Qualitative analysis of existing methods with and without the BoxMask module integrated into the ImageNet VID dataset under different scenarios. Clearly, our BoxMask module facilitates SELSA~\cite{wu2019sequence} and TROI~\cite{gong2021temporal} to alleviate misclassification and imprecise localization in case of rare pose ((a) turtle), motion blur, and part occlusion ((b) watercraft), respectively. Best view it on the screen and zoom in.}
\label{fig:qual_results}
\end{center}
\vspace{-20pt}
\end{figure}

\vspace{3pt}
\noindent \textbf{Visual Detection Results.} Figure~\ref{fig:qual_results} illustrates the detection results of two recent state-of-the-art methods integrated with our BoxMask module in odd and even rows, respectively. We can see that SELSA~\cite{wu2019sequence} yields false negatives (turtle as a background) and false positives (turtle as a bird) in the case of rare poses in (a). On the other hand, these false detections are reduced with the introduction of our BoxMask module. Similarly, in the case of motion blur and part occlusion, our method alleviates misclassification (watercraft as a car by TROI~\cite{gong2021temporal}) by learning fine-grained pixel-level temporal information. These results show that adopting the BoxMask module in region-based VOD methods introduces class-aware pixel-level feature aggregation across different video frames that facilitates VOD under challenging conditions. Refer to supplementary material for more qualitative analysis.

\begin{figure*}
\begin{center}
\includegraphics[width=15.3cm]{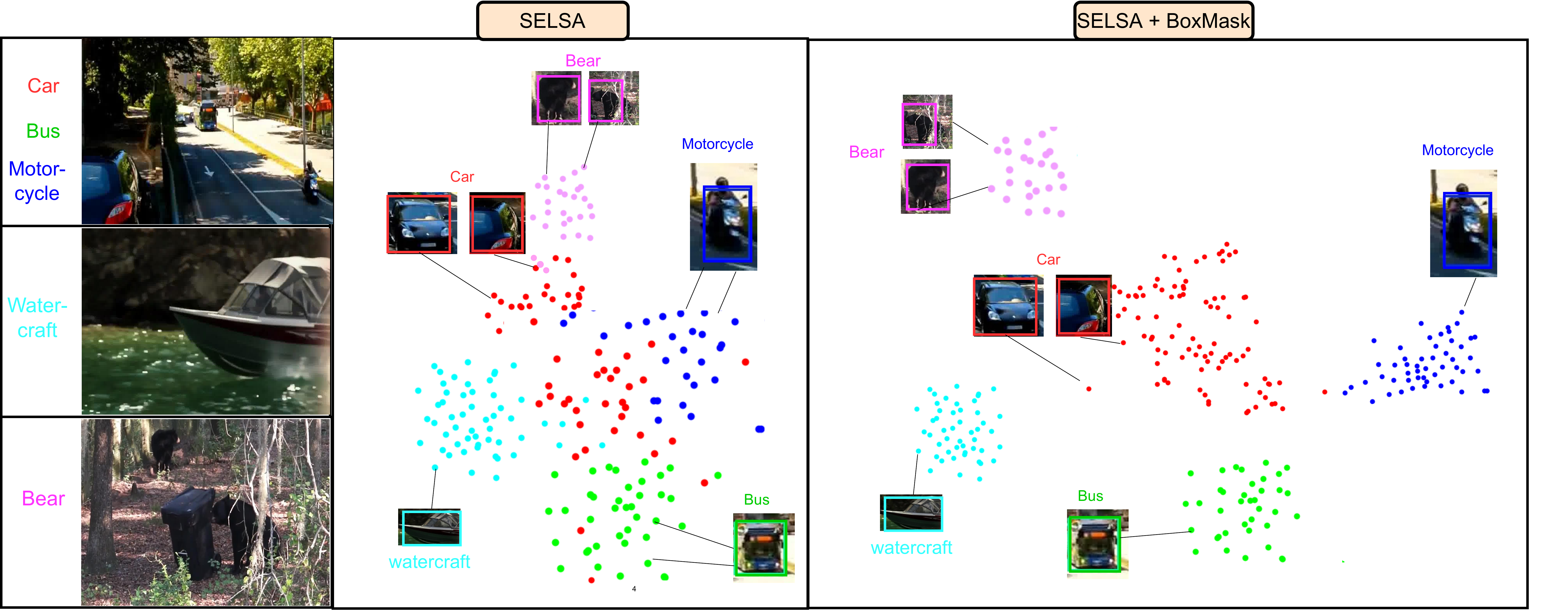}
\caption{t-SNE visualization of learned proposal features with and without our BoxMask module in SELSA~\cite{wu2019sequence}. With instance-level feature aggregation only in SELSA, proposals of objects with similar motion characteristics (\textit{Bus}, \textit{car}, and \textit{Watercraft}) mistakenly fall into each other's cluster. Our class-aware pixel-level learning in BoxMask introduces discriminative cues which alleviate this object confusion, as shown in SELSA+BoxMask. Best view in color. For the complete figure with all 30 categories, refer to supplementary material.}
\label{fig:tsne_vis}
\end{center}
\vspace{-20pt}
\end{figure*}

\vspace{3pt}
\noindent \textbf{Visual Proposal Feature Analysis.}
\label{sec:vis_features_anlysis}
Following~\cite{han2020mining}, we extract the learned proposal features before classification on the target frame and visualize them with t-SNE in Figure~\ref{fig:tsne_vis}. We can see that proposal features of SELSA misclassifies proposals into incorrect clusters. For instance, proposals of watercraft and a bus incorrectly fall into the cluster of a car due to similar appearance and motion characteristics. Alternatively, when BoxMask is integrated into SELSA~\cite{wu2019sequence}, we observe that proposal features of confusing object categories are clearly separated from each other. The main reason is that pixel-level feature aggregation enables the network to correctly distinguish proposals by decreasing the intra-class and increasing the inter-class variance. 

\begin{figure}
\begin{center}
\includegraphics[width=8.5cm]{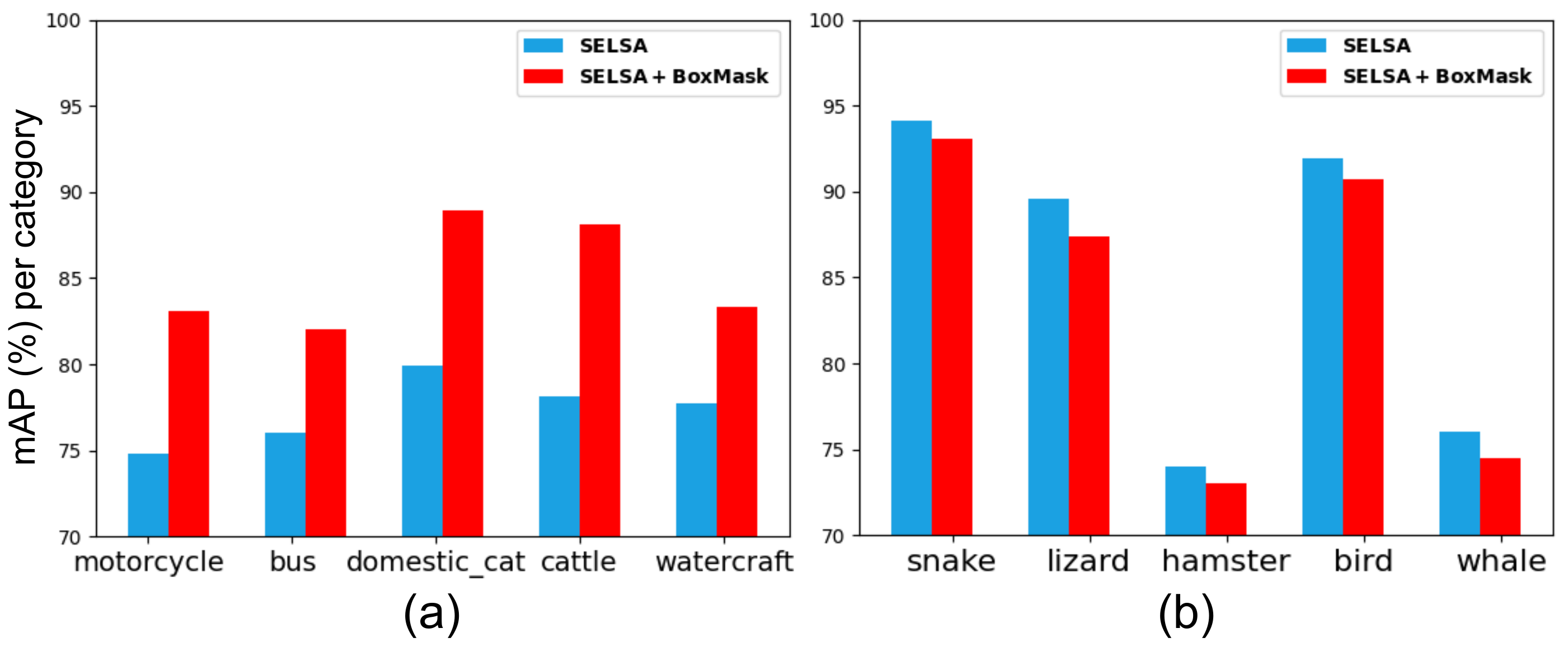}
\caption{Performance comparison in terms of mAP per category. Subplots (a) and (b) denote the five most improved and most dropped classes when the BoxMask is equipped in SELSA~\cite{wu2019sequence}.}
\label{fig:map_per_class}
\end{center}
\vspace{-20pt}
\end{figure}

\vspace{3pt}
\noindent \textbf{Analysis on Object Categories.}
Since our work mainly alleviates the object confusion in videos, we compare the performance in terms of mAP per category between the modern RoI-based VOD method~\cite{wu2019sequence} incorporated with and without our BoxMask module. We present the top 5 most improved classes and the top 5 most worsened categories in Figure~\ref{fig:map_per_class}. It is evident that the introduction of our pixel-level feature learning produces significant performance gains in {motorcycle}, {domestic\_cat}, and {cattle}.~The reason is that these objects have low inter-class variance due to similar appearance and motion characteristics.~The pixel-level learning in our BoxMask effectively tackles this challenge and improves overall performance, as illustrated in Figure~\ref{fig:qual_results}.

\subsection{Ablation Studies}
\begin{figure}
\begin{center}
\includegraphics[width=8cm]{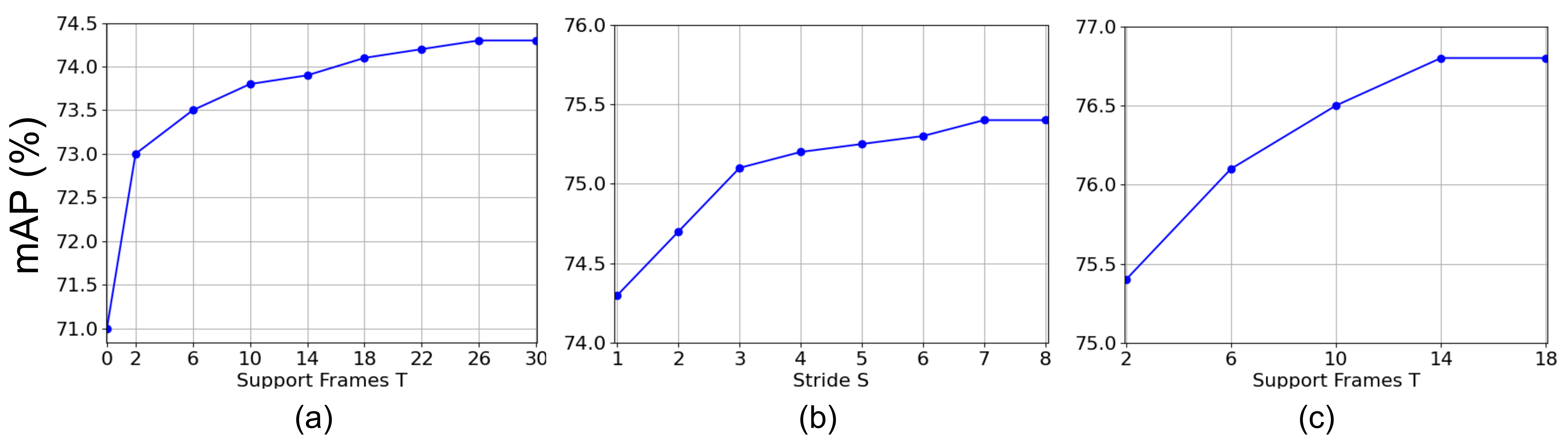}
\caption{Ablation studies for frame sampling methods. (a) Investigating the effect of the different number of frames by fixing frame stride to 1. (b) Examining the effect of different frame strides by fixing the number of support frames to 26. (c) Assessing the effect of a different number of support frames sampled over the complete video.}
\label{fig:frame_sampling}
\end{center}
\vspace{-20pt}
\end{figure}

\noindent \textbf{ Sampling Support Frames.} We follow the spirits of~\cite{wu2019sequence} and~\cite{gong2021temporal} to analyse the influence of number of frames and sampling strides during testing. Moreover, we examine the number of support frames sampled over an entire video. Figure~\ref{fig:frame_sampling}\color{red}{(a)} \color{black} exhibits the influence of an increasing number of support frames \textit{T}. We start with a singe-frame detector by setting the frame stride \textit{S} to 1.~The mAP improves with the increasing number of frames, and it tends to stabilize at 74.4 mAP at $T=26$. Later, we set \textit{T} to 26 and start increasing the frame stride \textit{S}. As illustrated in Figure~\ref{fig:frame_sampling}\color{red}{(b)}\color{black}, the mAP consistently improves with rising stride and eventually settles at $S=7$. Finally, to exploit the whole video information, we make \textit{S} adaptive to the length of the video corresponding to the number of support frames \textit{T}. Figure~\ref{fig:frame_sampling}\color{red}{(c)}\color{black}~shows that by leveraging only 2 support frames sampled over the entire video, mAP of 75.4 is achieved, surpassing the mAP on 26 succeeding support frames. The performance further boosts with the rise in the number of support frames and finally stabilizes at 14. We use the uniform sampling method with $T=14$ in all the experiments unless stated otherwise.

\setlength{\tabcolsep}{2pt}
\begin{table}
\begin{center}
\footnotesize
\begin{tabular}{lllll}
\toprule
\textbf{Methods} & \textbf{AP\textsubscript{0.5}}(\%) & \textbf{AP\textsubscript{0.75}}(\%) & \textbf{AP\textsubscript{0.5:0.95}}(\%) &  \textbf{Runtime(FPS)}\\
\noalign{\smallskip}
\hline
\noalign{\smallskip}
DFF~\cite{zhu2017deep}  & 70.3 & 45.7 & 42.7 & 58.8 \\
FGFA~\cite{zhu2017flow}  & 74.7 & 52.0 & 47.1 & 19.3 \\
SELSA~\cite{wu2019sequence}  & 78.4 & 52.5 & 48.6 & 13.9 \\
TROI~\cite{gong2021temporal}  & 78.9 & 52.8 & 48.6 & 7.3 \\
\hline

DFF + BoxMask & 71.3$_{\uparrow}$\textsubscript{1.0} & 47.7$_{\uparrow}$\textsubscript{2.0} & 44.6$_{\uparrow}$\textsubscript{1.9} & 51.8$_{\downarrow}$\textsubscript{8.0} \\
FGFA + BoxMask & 75.6$_{\uparrow}$\textsubscript{0.9} & 54.2$_{\uparrow}$\textsubscript{2.2} & 49.4$_{\uparrow}$\textsubscript{2.3} & 17.3$_{\downarrow}$\textsubscript{2.0} \\
SELSA + BoxMask & 79.5$_{\uparrow}$\textsubscript{1.1} & 55.7$_{\uparrow}$\textsubscript{3.2} & 50.6$_{\uparrow}$\textsubscript{2.0} & 13.0$_{\downarrow}$\textsubscript{0.9} \\
TROI + BoxMask & 80.7$_{\uparrow}$\textsubscript{1.8} & 57.8$_{\uparrow}$\textsubscript{5.0}& 51.8$_{\uparrow}$\textsubscript{3.2} & 7.2$_{\downarrow}$\textsubscript{0.1} \\
\bottomrule\\
\end{tabular}
\caption{Tradeoff between effectiveness and efficiency of the proposed BoxMask using ResNet-50 as the backbone network. The run time is tested on a single DGX A100 GPU.}
\label{table:sota_resnet-50}
\end{center}
\vspace{-20pt}
\end{table}
\setlength{\tabcolsep}{1.4pt}

\vspace{3pt}
\noindent \textbf{Effectiveness of BoxMask.}
To investigate the flexibility and effectiveness of our method, we reproduce 4 existing RoI-based VOD methods~\cite{gong2021temporal, wu2019sequence, zhu2017flow, zhu2017deep} and incorporate our BoxMask module. Table~\ref{table:sota_resnet-50} summarizes the trade-off between the effectiveness and efficiency of the proposed BoxMask module. Looking at Table~\ref{table:sota_resnet-50}, we observe regular and substantial improvements in all 4 methods when equipped with our simple yet effective BoxMask module. Adopting BoxMask in TROI, we achieve a mAP@0.5 of 80.7(\%), the new state-of-the-art result on ResNet-50. Similarly, on an increasing IoU threshold of 0.75, we notice a significant gain of 5 points in mAP when BoxMask is equipped with TROI. This reflects that our pixel-level learning not only alleviates object confusion but also yields high-quality predictions. 

\vspace{3pt}
\noindent \textbf{Effect on increasing convolutions.}
\label{sec:ablation_num_conv}
We investigate the design of our BoxMask network prior to pixel-wise class prediction. Specifically, for the instance feature extraction head, we study the impact of an increasing number of $3 \times 3$ convolutional layers, N\textsubscript{c}. As shown in Table~\ref{table:increasing_conv}, the mAP declines with the rise in N\textsubscript{c}. We argue that there are two main reasons for such behaviour.
First, since our BoxMask aims to predict a mask of a rectangular target object, an increasing number of convolutional layers introduce unnecessary complex parameters that lead to overfitting. Second, given that our BoxMask is supervised on bounding box annotations (containing object and background), increasing the size of N\textsubscript{c} allows the network to learn needless high-level features, which causes object confusion.

\vspace{3pt}
\noindent \textbf{Size of RoI Features.}
As explained in Section~\ref{sec:roi_feature}, we perform Temporal RoiAlign to extract RoI features with spatial resolution of $ 7 \times 7$. The RoI features are then upsampled to size $14 \times 14$. Furthermore, we investigate different resolution settings for RoI and upsampled features for completeness. As summarized in Table~\ref{table:roi_size}, we accomplish an optimal trade-off between performance and efficiency upon setting the RoI and upsampled features size as 7 and 14, respectively.


\begin{minipage}{\linewidth}
  \begin{minipage}{0.4\linewidth}
    \vspace{-10pt}
    \begin{table}[H]
      \begin{center}
        {\small{
            \begin{tabular}{c|ccc}
            \toprule
            (N\textsubscript{c}) & mAP(\%) & FPS & Params\\
            \noalign{\smallskip}
            \midrule
            \noalign{\smallskip}
            1 & \textbf{80.7} & 7.2 & 0.8\\
            2 & 80.5 & 6.8& 1.5\\
            3 & 80.4 & 6.6& 2.1\\
            4 & 80.2 & 6.2& 2.6\\
            \bottomrule
            \end{tabular}
            }}
        \end{center}
    \caption{Effect on increasing number of convolutional layers (N\textsubscript{c}) in BoxMask. Params represents number of parameters $\times 10^6$.}
    \label{table:increasing_conv}
    \end{table}
  \end{minipage}
  \hspace{0.05\linewidth}
  \begin{minipage}{0.4\linewidth}
  \vspace{-50pt}
\begin{table}[H]
  \begin{center}
    {\small{
        \begin{tabular}{cc|cc}
            \toprule
            RoI & Upsample & mAP(\%) & FPS\\
            \noalign{\smallskip}
            \midrule
            \noalign{\smallskip}
            7 & 7 & 78.3 & 7.4\\
            7 & 14 & \textbf{80.7} & 7.2\\
            7 & 28 & 79.4 & 6.5\\
            14 & 14 & 80.7 & 5.2 \\
            \bottomrule
        \end{tabular}
        }}
    \end{center}
        \caption{Effect on increasing size of RoI Features.}
\label{table:roi_size}
\end{table}
  \end{minipage}
\end{minipage}

\vspace{3pt}
\noindent \textbf{Computational Analysis.}
For brevity,~we present the real-time performance of our BoxMask module in Table~\ref{table:sota_resnet-50}.~We can observe that the speed of SELSA and SELSA+BoxMask are 13.9 and 13.0 FPS on a single DGX A100 GPU, respectively. Moreover, when our method is adopted in TROI~\cite{gong2021temporal}, the speed drops by 0.1 FPS while achieving an mAP gain of 1.8 points.~This demonstrates that the BoxMask module brings significant performance gains with a negligible increase in computation.

\subsection{Additional Experiments on EPIC KITCHENS}

\vspace{3pt}
\noindent \textbf{Dataset and Implementation Details.}
Along with ImageNet VID dataset, we evaluate our method on a far-more challenging EPIC KITCHENS dataset~\cite{damen2018scaling}.~The VOD task in this dataset comprises 32 unique kitchens, including 290 classes.~We employ 272 video sequences captured in 28 kitchens for training, whereas, for evaluation, 106 sequences are collected in the same 28 Kitchens (S1), and 54 sequences are gathered from other 4 unseen kitchens (S2). For direct comparison with prior works~\cite{gong2021temporal, wu2019sequence}, we adopt identical implementation settings explained in~\cite{wu2019sequence}.

\vspace{3pt}
\noindent \textbf{Performance Analysis.}
We reimplement prior works~\cite{wu2019sequence, gong2021temporal} on the EPIC KITCHENS dataset and evaluate the results on an IoU threshold of 0.5 and 0.75. As summarized in Table~\ref{table:epic_kitchen_result}, we observe consistent and significant performance gains when the proposed BoxMask is equipped in SELSA~\cite{wu2019sequence} and TROI~\cite{gong2021temporal}. It is important to emphasize that on a higher IoU threshold of 0.75, our BoxMask further improves the mAP to ($4.5 / 3.6$) points for SELSA and ($5.2/4.4$) points for TROI for Seen/Unseen splits. This establishes that incorporating our simple BoxMask module in region-based detectors can boost the performance of VOD even on complex datasets.

\begin{table}
  \begin{center}
    {\footnotesize{
        \begin{tabular}{lll|ll}
        \toprule
        Methods & AP\textsubscript{0.50}(S1) & AP\textsubscript{0.75}(S1) &AP\textsubscript{0.50}(S2) & AP\textsubscript{0.75}(S2) \\
        \noalign{\smallskip}
        \hline
        \noalign{\smallskip}
        SELSA~\cite{wu2019sequence}  & 38.8  & 10.2 & 36.7 & 9.2 \\
        TROI~\cite{gong2021temporal}  & 42.2 & 13.3 & 39.6 & 11.3 \\
        \hline
        SELSA + BoxMask & 40.7$_{\uparrow}$\textsubscript{1.9} & 14.7$_{\uparrow}$\textsubscript{4.5} & 38.1$_{\uparrow}$\textsubscript{1.4} & 12.8$_{\uparrow}$\textsubscript{3.6} \\
        TROI + BoxMask & 44.3$_{\uparrow}$\textsubscript{2.1} & 18.5$_{\uparrow}$\textsubscript{5.2}& 41.3$_{\uparrow}$\textsubscript{1.7} & 15.7$_{\uparrow}$\textsubscript{4.4} \\
        \bottomrule
        \end{tabular}
        }}
    \end{center}
    \caption{Performance comparison without and with the BoxMask module in previous state-of-the-art methods on EPIC KITCHENS test set. S1 and S2 represent Seen and Unseen splits, respectively.}
\label{table:epic_kitchen_result}
\vspace{-10pt}
\end{table}

\subsection{Limitations}
Albeit integrating our proposed module in VOD systems substantially improves detections, we notice that the performance drops for some object categories, as illustrated in Figure~\ref{fig:map_per_class}\color{red}{(b)}\color{black}. We observe that our method yields false negatives (confusing objects with background) and false positives (confusing background with an object class). Such behaviour is due to the supervision from faulty object masks with no information on object boundaries. Therefore, our BoxMask network treats the background part in the bounding box as an object mask by learning class-aware pixel-level information. Thus, applying methods like GrabCut~\cite{rother2004grabcut} and MCG~\cite{arbelaez2014multiscale} on bounding boxes to reduce the background content in object masks is one possible way to tackle this problem. Moreover, learning the refined instance mask from the coarse BoxMask in a weakly supervised manner as done in still images~\cite{lan2021discobox, tian2021boxinst} will introduce refined object boundaries, alleviating confusion between foreground and background pixels.

\section{Conclusion}
In this paper, we address the crucial problem of object confusion that limits the upper bound of video object detection models and present a simple yet effective BoxMask module as a remedy.~Our method introduces class-aware pixel-level information that brings crucial discriminative indicators that enhance classification and localization.~The proposed module is conceptually simple and can be applied to any region-based detection method to boost performance. Extensive experiments on ImageNet VID and EPIC KITCHENS datasets demonstrate that the introduction of our proposed method brings consistent and significant performance gains in recent video object detection methods.

{\small
\bibliographystyle{ieee_fullname}
\bibliography{egbib}
}

\end{document}